\documentclass[runningheads]{llncs}

\usepackage{eccv}

\usepackage{eccvabbrv}

\usepackage{graphicx}
\usepackage{booktabs}
\usepackage[accsupp]{axessibility}  

\usepackage{hyperref}

\usepackage{orcidlink}

\newcommand{\centered}[1]{\begin{tabular}{l} #1 \end{tabular}}

\begin{document}

\renewcommand{\topfraction}{.8}
\renewcommand{\floatpagefraction}{.8}

\title{ExeChecker: Where Did I Go Wrong?} 

\titlerunning{ExeChecker}

\author{Yiwen Gu\inst{1}\orcidlink{0000-0002-2437-0343} \and
Mahir Patel\inst{1}\orcidlink{0000-0002-3370-4595} \and
Margrit Betke\inst{1}\orcidlink{0000-0002-4491-6868}}

\authorrunning{Y.~Gu \etal}

\institute{Boston University, Boston MA 02125, USA \\
\email{\{yiweng, mahirp, betke\}@bu.edu}}

\maketitle

\begin{abstract}
  In this paper, we present a contrastive learning based framework, \textbf{ExeChecker}, for the interpretation of rehabilitation exercises. Our work builds upon state-of-the-art advances in the area of human pose estimation, graph-attention neural networks, and transformer interpretablity. The downstream task is to assist rehabilitation by providing informative feedback to users while they are performing prescribed exercises.  We utilize a contrastive learning strategy during training. Given a tuple of correctly and incorrectly executed exercises, our model is able to identify and highlight those joints that are involved in an incorrect movement and thus require the user's attention. 
  We collected an in-house dataset, \textbf{ExeCheck}, with paired recordings of both correct and incorrect execution of exercises.  In our experiments, we tested our method on this dataset as well as the UI-PRMD dataset and found ExeCheck outperformed the baseline method using pairwise sequence alignment in identifying joints of physical relevance in rehabilitation exercises.
  
\end{abstract}

\section{Introduction}
\label{sec:intro}

\begin{figure}[t]
  \centering
  \includegraphics[width=0.9\linewidth]{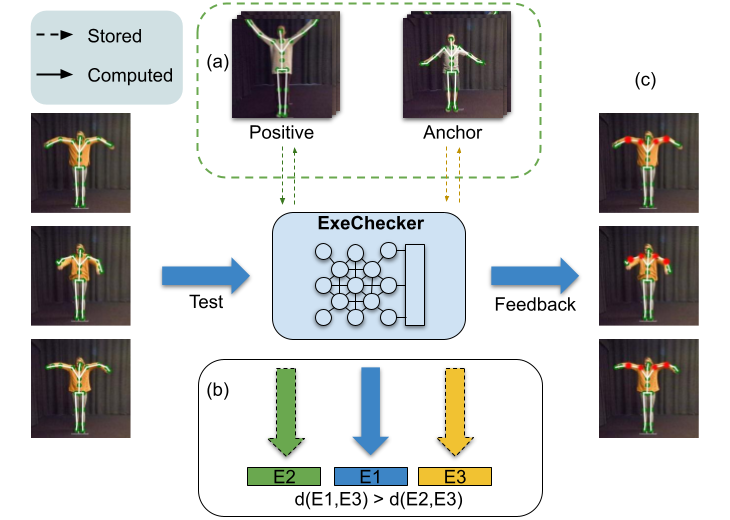}
  \caption{ExeChecker interprets the movement of skeleton joints extracted from RGB video and provides feedback to the user about which joints to pay attention to. (a) We first generate a database by pre-computing and storing the embeddings for the positive and anchor sequences. (b) During inference, we compute embedding E1 for the input sequence $M$ and test it against the positive (E2) and anchor (E3) pair. (c) By using the attention score from our graph attention network, we highlight the joints (red) that require adjustments. }
  \label{fig:what-is-this-paper-about}
\end{figure}

Rehabilitation exercises help individuals recover from injuries, surgeries, or chron\-ic conditions. In the clinical setting, physical therapists consider the physical conditions of the patients and prescribe them different sets of exercises or the same exercise but with different standards. 
In most cases, patients learn the prescribed exercises during their clinic visits but then perform them regularly in their homes. In the absence of a therapist, patients do not receive adequate guidance and instructive feedback. This may lead to a lack of motivation and incorrect exercise performance, negatively impacting the patient's health. To help patients better adhere to the prescribed exercises, researchers build home-based systems that take the patient's movements as input, analyze their performance, and automatically provide feedback~\cite{saraee2019exercisecheck, pandit2019exercisecheck}.
However, most feedback generated by those systems is too general ("excellent/ good/ needs more work") or too obscure 
(some score). Such feedback does not provide information regarding which movements the patients did incorrectly and hence need attention. In this paper, we propose a framework, \textbf{ExeChecker}, that tells incorrect from the correct movement for a set of rehabilitation exercises and points out the joints that cause the incorrect performance (Fig.~\ref{fig:what-is-this-paper-about}).

\textbf{ExeChecker} takes skeleton-based human poses, leverages the existing spatial-temporal graph attention transformer, and adopts a triplet learning strategy. During the training, the model is forced to minimize the distances between the positive pairs (consisting of the sequences of correctly conducted exercises) and maximize the distances of negative pairs (consisting of correctly and incorrectly conducted exercises) in the embedding space.  As a result, the learned embedding focuses on the dissimilarities between the incorrect and correct exercises.

We conducted extensive experiments on two datasets, UI-PRMD~\cite{vakanski2018data} and our in-house collected dataset \textbf{ExeCheck}. Both datasets feature common rehabilitation exercises and paired recordings of both correct and incorrect performance. 

Our contributions are summarized as follows:
\begin{itemize}
\item We collaborated with a physical therapist who specializes in exercise rehabilitation and collected an in-house dataset that includes ten frequently prescribed exercises for patients with Parkinson's disease. For each exercise, we paired the correct performance with incorrect ones that represent the most common mistakes.
\item We propose a novel strategy that leverages similarity learning to identify joints that result in incorrect movement and require more attention from the patients. 
The identification of those joints is self-supervised. 
We compose triplets of correctly and incorrectly performed exercises and force the model to learn embeddings that maximize the differences between the correctly and incorrectly performed exercises while minimizing the distances among the correctly performed ones. 
\item We invented a scoring system to evaluate the performance of models in terms of correctly identifying the joints-of-attention (JoA), which are the joints with physical relevance leading to incorrectness when doing an exercise. We provide informative visual feedback that highlights those joints for users to see "Where did I go wrong?".
\end{itemize}

The collected dataset, the 3D joint positions, and the attention scores per joint will be publicly available at {\tt \small http://www.cs.bu.edu/faculty/betke/ExeChecker}.

\section{Related Work}

Movement assessment, also called action quality assessment (AQA), refers to the evaluation and analysis of physical movements or actions to determine their quality, efficiency, and effectiveness. This assessment is useful in various fields, including sports, physical therapy, and robotics. 
The assessment could be classifying the form of the movement~\cite{Ogata_2019_CVPR_Workshops, parmar2022domain, gao2023automating, wang2022skeleton}, or providing a final score representing the quality of the movements~\cite{deb2022graph, parmar2017learning, parmar2019and, pan2019action, tang2020uncertainty, Gu_2019_ICCV, sardari2020vi, liao2020deep}. 

In the field of sports, the quality of an action may be assessed by a judge or a panel of judges. 
Parmar \etal~\cite{parmar2017learning} constructed a dataset on Olympic events from YouTube videos and proposed the use of spatio-temporal features from 3D convolutional neural networks (C3D) for predicting quality assessment scores \cite{parmar2017learning}. 
Their work was extended by considering assessment as a multitask~\cite{parmar2019and}; 
besides computing an action quality score, the extended model also provides a summary of action events (e.g., number somersaults) and generates captions.
Pan~\etal\cite{pan2019action} combined 
features extracted from local image patches around joints with pose information by building temporal and spatial joint-relation graphs to assess action performance. 
Tang~\etal~\cite{tang2020uncertainty} proposed an uncertainty-aware score-distribution learning approach to address the intrinsic ambiguity in the score labels caused by different judges.

When action assessment scores are not readily available, movements can also be assessed by classifying the form, \ie, whether it is correct or what error category it belongs to.
Ogata~\etal~\cite{Ogata_2019_CVPR_Workshops} proposed a ResNet-based model analyzing temporal distance matrices for the classification of squat exercises.
Parmar \etal~\cite{parmar2022domain} introduced a Fitness-AQA dataset on weight lifting videos in gyms and proposed two self-supervised approaches, pose contrastive learning and motion disentangling, to assess errors in the execution of the exercises.

Contrastive learning has also been used to improve classification accuracy for the diagnosis of motor dysfunction in infants.  
Gao~\etal~\cite{gao2023automating} presented a motor assessment model to identify "fidgety movements" (FM). In their model, they included a branch trained with a triplet loss to differentiate  positive (FM) and negative (non-FM) representations.  In our work, we also designed a triplet loss, here to distinguish positive and negative examples of correct exercising. 

In the field of rehabilitation, exercise performance is typically assessed by physical therapists.  Data collection can be conducted in controlled environments, \eg, a laboratory or clinic, using cameras and/or maker-based or markerless motion capture systems.
The two datasets most widely used for research in exercise assessment are {\it KInematic Assessment of MOvement and Clinical Scores for Remote Monitoring of Physical REhabilitation} (KIMORE)~\cite{Kimore2019} and the {\it University of Idaho-Physical Rehabilitation Movement Data} (UI-PRMD)~\cite{vakanski2018data}. Both datasets provide skeletal joint data.  The KIMORE dataset includes RGB-D exercise recordings of a group of 34 participants who suffer from chronic motor disabilities and a group of 44 healthy participants. KIMORE provides score annotations where the quality of the exercise performance is evaluated by physical therapists.  The UI-PRMD involves joint data of 10 healthy participants conducting physical therapy movements in correct and incorrect forms. UI-PRMD does not include score annotations. 
Instead, in their subsequent work~\cite{liao2020deep}, the researcher group proposed a scoring function that maps the values of the performance metric into a movement quality score in the range between 0 and 1. They adopted the log-likelihood of a Gaussian Mixture Model as the performance metric for evaluating the movements~\cite{liao2020deep}.
Other works on rehabilitation exercise assessment exploited the inherent spatial and temporal relations in the skeletal joints and developed graph-based neural networks for quality assessment. 
More generally, in the field of human action recognition, various strategies that incorporate spatial and temporal information into graph convolutional networks or graph attention transformers have been investigated 
(survey~\cite{xin2023transformer}). Correspondingly, rehabilitation-orientated graph-based models with attention mechanisms have been proposed~\cite{deb2022graph,mourchid2023d,reby2023graph} and evaluated using UI-PRMD and/or KIMORE as benchmark datasets.
In our study, we conducted experiments with UI-PRMD, in addition to our ExeCheck dataset, for their paired collection of correctly and incorrectly performed exercises.

\section{Method}

\begin{figure}[tb]
  \centering
  \includegraphics[height=7.5cm]{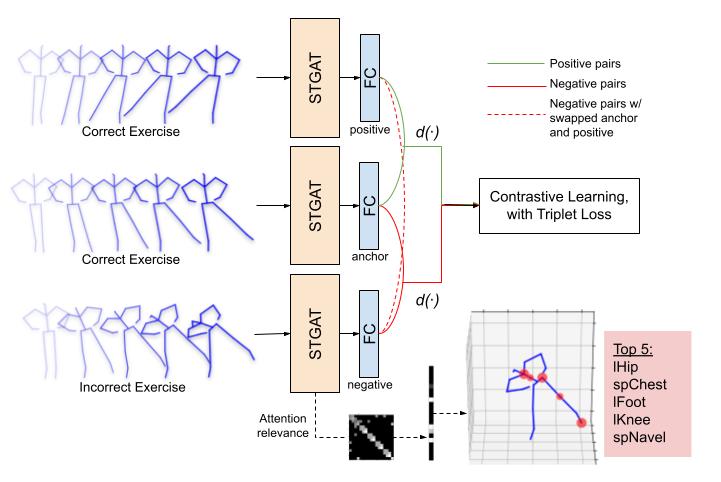}
  \caption{The ExeChecker framework: The top five body joints are identified that distinguish correct and incorrect exercise movements using contrastive learning.  }
  \label{fig:model_arch}
\end{figure}

Given an exercise movement 
$M = [\mathbf{x}_1, \mathbf{x}_2, ..., \mathbf{x}_T]$ where $\mathbf{x}_t \in \mathbb{R}^{N \times 3} $ is the human pose at time $t$ and is represented by $N$ skeletal joints in 3D positions. ExeChecker
1) classifies $M$ as either correctly or incorrectly performed exercise; and 2) if $M$ is of the incorrect form, ExeChecker outputs the body joints that contribute to the incorrectness.

The architecture of ExeChecker is a triplet network, shown in Fig.~\ref{fig:model_arch}. It leverages the design of a state-of-the-art spatial-temporal graph transformer for feature extraction, \ie, embedding generation (Section~\ref{sec:stgat}), and then uses the triplet training strategy (Section~\ref{sec:triplet}) to maximize the distance in the embedding space between correct and incorrect movements.  After being trained, the output of the attention map from the feature extraction module is used to identify the Joint of Attention if the input sequence is an incorrect movement.

\subsection{Feature Extraction with STGAT}
\label{sec:stgat}

Spatial-temporal graph transformers are designed for the task of skeleton-based action recognition. 
Since the core of human action recognition is to obtain discriminative feature representation~\cite{presti20163d}, it aligns with our goal of finding discriminating feature embeddings that separate the correct and incorrect exercise movements. A variety of graph transformers has been proposed over the past years, and we recommend a recent survey~\cite{xin2023transformer} for readers interested in this topic. 
In our implementation, we adopted the same STGAT architecture as described in the work of Hu \etal~\cite{hu2022spatial} as our feature extractor. Compared to prior works, they added a local spatial-temporal feature aggregator, which enables the model to better capture local cross-spacetime joint relationships and makes it more sensitive to similar action pairs. Briefly, at a given timestamp $t$, each joint node in pose $\mathbf{x}_t$ is not only connected to all the other joints (plus itself) in the current pose but also every joint in $\tau$ neighboring poses. As a result, the cross-spacetime adjacency matrix at time $t$ is $A_{\tau}^t$ is ${\tau \times N \times N}$ in dimension. By incorporating $A_{\tau}^t$ into graph operations and applying the multi-head self-attention, 
the output of STGAT can be obtained as 
\begin{equation}
    Y^t = \sigma( \frac{1}{H} \sum_{h=1}^H W_h X_{\tau}^t A_{\tau, h}^t)
\end{equation}
where $X_{\tau}$ is a sequence of $\tau$ poses centered at timestamp $t$ and $H$ is the number of attention heads. For our experiments, we set $\tau=3$ and $H=8$.

\subsection{Triplets Training}
\label{sec:triplet}

Triplet networks have proven to be suitable for applications where fine-grained distinction between classes is important, such as face recognition~\cite{schroff2015facenet,hermans2017defense} and fine-grained image retrieval~\cite{yao2020adaptive, D'Innocente_2021_CVPR}. 
Triplet loss  \cite{Wang_2014_CVPR}  is a triple-based ranking loss function defined as 
\begin{equation}
\label{eq:tripletLoss_m}
    \mathcal{L} (a,p,n) = \max(0, d(a,p) - d(a,n) + \mu),
\end{equation}
where $a, p$, and $n$ denote the anchor, positive and negative samples respectively, $d(\cdot)$ is a distance function and $\mu$ is a margin parameter.
This ranking loss measures the violation of the ranking order of the embedded features inside the triplet. Pairwise distance metrics such as Euclidean distance or cosine similarity are valid distance functions.

The margin~$\mu$ in ~\Cref{eq:tripletLoss_m} introduces an extra hyperparameter during the triplet training. With a margin too small, the network can quickly learn to satisfy the triplet constraint, leading to a triplet loss that drops to zero early in training. As a result, the learned embeddings may not be well separated, and the network fails to learn discriminative features for different classes. On the other hand, if the margin is too large, the network may find it hard to satisfy the triplet constraint and struggle to converge. 
In a separate study~\cite{hoffer2015deep}, a softmax function is applied to the distance pairs, creating a ratio measure. The proposed triplet ratio loss can be formulated as 
\begin{equation}
\label{eq:tripletLoss_r}
    \mathcal{L} (a,p,n) = (\frac{e^{d(a,p)}}{e^{d(a,p)}+e^{d(a,n)}})^2. 
\end{equation}

In our experiments, we followed the in-triplet hard negative mining with anchor swap strategy~\cite{balntas2016learning}. Briefly, if $d(p,n) < d(a,n)$, we swap anchor and positive samples. This ensures that the hardest negative inside the triplet is used for backpropagation. We initially used the margin loss (\Cref{eq:tripletLoss_m}) for training the triplets. The training was unstable, and the loss did not converge despite trying various margin values. Consequently, we use the ratio loss (\Cref{eq:tripletLoss_r}) for our experiments.

\subsection{Visual Feedback on Incorrect Joint Movement }

In order to evaluate if the model correctly identifies the joints of attention, we design a scoring system described as follows. 

Given an exercise $e$ and the corresponding set of Joints of Attention, $JoA_e$, the number of elements inside denoted as $N_e$ ($N_e = |JoA|$) varies depending on the $e$.  Denote the complete joints set that represents a skeletal pose as $J= \{J_i\}, (i=1,2,..,N) $, we have $JoA \subset J$. For each exercise, the $i$th joint is assigned a score~$s_i$ such that 
\begin{equation}
    s_i =\begin{cases}
        1 & \text{if $J_i \in JoA$ }\\
        0 & \text{otherwise}.
    \end{cases}
\end{equation}
Our method computes a score for each joint. We first min-max normalize the scores to $[0,1]$. Let $\hat{s_i}$ denote the computed score for joint $J_i$, we calculate score of JoA as 
\begin{equation}
    S_{JoA} = \frac{1}{N_e} \sum_{i=1}^N s_i \times \hat{s_i},
 \end{equation}
where $S_{JoA} \in [0,1]$.

\section{ExeCheck Dataset}
\label{sec:exec_ds}
The ExeCheck consists of RGB-D videos of 10 rehabilitation exercises performed by 7 healthy subjects. Each exercise has a paired performance in both correct and incorrect forms by the same subject with 5 movement repetitions.

\textbf{Data Collection.} 
We collaborated with a physical therapist from a rehabilitation science program from an accredited university, who selected ten routinely prescribed rehabilitation exercises for patients with Parkinson's disease and demonstrated to us both their correct and incorrect performance (Table~\ref{tab:dataset_exec}). Specifically, the incorrect form highlights the most common mistakes observed by therapists.  A total of 7 healthy subjects were recruited, and their consent for participating in the study was obtained according to a protocol approved by the university's Institutional Review Board.  

Each subject performed 10 exercises weaing relatively tight-fitting clothes.  They performed each exercise five times, first in a correct way and then in an incorrect way, by purposely making at least one of the common mistakes listed in Table~\ref{tab:dataset_exec}.
 The data was collected in a lab setting using the MS Azure Kinect sensor. We recorded RGB-D videos for each exercise at a framerate of 30~fps. 

\textbf{Data Processing and Annotation.} 
We processed the recordings offline via the Azure Kinect Body Tracking SDK~\cite{azure_sdk} to extract skeletal joint information. For each video frame, we obtained the positions and orientations of 32 joints. We used the 3D coordinates of 21 of them for our experiments. The chosen joints include 17 joints compatible with the Human3.6M dataset format, plus the left and right hands and feet. We annotated each exercise with the joints of attention, where the correct exercise (CE) and incorrect exercise (IE) share the same set. We further segmented the repetitions of each exercise manually.

We will release our dataset, which includes both the original videos and processed joint files, as well as the repetition annotations.

\begin{table}
    \centering
    \caption{Joints of Attention in the ExeCheck dataset identified by our model. The letter in the brackets indicates the participant's side (L: left, R: right) that is actively engaged in the exercise. If not indicated, both sides are engaged. For incorrectly performed exercises, we asked the participants to err on a specific side and the corresponding joints of attention are denoted by $^*$.}
    \label{tab:dataset_exec}
    \begin{small}
    \begin{tabular}{p{0.25\linewidth}|p{0.45\linewidth}|p{0.27\linewidth}} \hline
         \centered{\textbf{Exercise} \\ \textbf{Name}} 
         & \centered{\textbf{Included Common Mistakes}} 
         & \centered{\textbf{Joints of Attention}} \\ 
         \hline
         \hline
         \centered{Arm Circle \\ (0-AC) }
         & \centered{Hunching shoulders \\ Not keeping arms extended straight }
         & \centered{lShoulder, rShoulder, \\ lElbow, rElbow} \\[1ex]
         \hline
         \centered{Forward Lunge\\ (1-FL) [L]}
         & \centered{Leaning forward \\ Letting the knee go past the toe \\ Slamming right knee on the ground} 
         & \centered{spChest, \\ lFoot, \\ rKnee} \\[1ex]
         \hline
         \centered{High Knee Raise\\ (2-HKR) [L]}
         & \centered{Unstable upper body and wobbing \\ Not raising the knee high enough \\ Slamming foot (dropping the leg too \\quickly)}
         & \centered{spNavel, spChest, \\ lKnee, \\ lFoot} \\
         \hline
         \centered{Hip Abduction\\ (3-HA) [L] }
         & \centered{Tilting the upper body to the \\opposite side for compensation} 
         & \centered{lHip, \\ spNavel, spChest} \\
         \hline
         \centered{Leg Extension\\ (4-LE) [L]} 
         & \centered{Leaning the upper body forward \\ Not fully extending the leg} 
         & \centered{spNavel, spChest, \\ lHip, lKnee \\ } \\
         \hline
         \centered{Shoulder \\Abduction \\ (5-SA)} 
         & \centered{Lifting the arms unequally and \\tilting the head for compensation \\ Hunching shoulders} 
         & \centered{rShoulder$^*$, rElbow$^*$, \\ neck \\ lShoulder} \\
         \hline
         \centered{Shoulder \\External Rotation \\ (6-SER) [L]}
         & \centered{Not keeping the elbow close to \\the body} 
         & \centered{lShoulder, lElbow }\\
         \hline
         \centered{Shoulder Flexion \\ (7-SF)} 
         & \centered{Not lifting arms equally high enough} 
         & \centered{lShoulder$^*$, lElbow$^*$} \\
         \hline
         \centered{Side Step Squat \\ (8-SSS) [L]} 
         & \centered{Rotating the upper body \\ Leaning the upper body forward }
         & \centered{lHip, rHip, \\ spNavel, spChest} \\
         \hline
         \centered{Squat \\ (9-Sq) } 
         & \centered{Bending knees inward (valgus knees) \\ Not keeping the back upright and \\straight} 
         & \centered{lKnee, rKnee, \\ spChest, spNavel}\\
         \hline
    \end{tabular}
    \end{small}
    \vspace{0.2cm}
\end{table}

\section{Experiments and Results}
\label{sec:exp}

\subsection{Datasets}
\label{sec:datasets}
We conducted experiments on two datasets, our \textbf{ExeCheck} dataset (see~\Cref{sec:exec_ds}) and the \textbf{UI-PRMD} dataset~\cite{vakanski2018data}.

The original UI-PRMD dataset includes joint positions of 10 persons performing 10 exercises, each with 2 sequences for correct and incorrect movements.  In some sequences in the UI-PRMD dataset, the published joint positions, especially those in the lower body, model unrealistic movements and suffer from unstableness. To circumvent this issue, we re-processed the original RGB videos using the state-of-the-art off-the-shelf model for human pose estimation PoseformerV2~\cite{zhao2023poseformerv2}. 
We also manually segmented the repetitions of each exercise based on the video time stamps.
The extracted human poses have 17 joints following the Human3.6M dataset format.
We further annotated the corresponding JoAs for each exercise based on the description provided in the UI-PRMD paper. Their descriptions of the non-optimal movements, \ie, the incorrect exercises, point out joints that need to be attended to. Moreover, their exercise set overlaps with ours, so some knowledge for JoA could be transferred. The difference mainly lies in the different joint sets. Our ExeCheck dataset has two spine joints (spine navel and spine chest) while the Human3.6M format has only one (torso). As an alternative, if, for an exercise, the incorrect version is described as "Subject unable to maintain upright trunk posture \ldots" we include 'torso' and 'neck' in the JoA set of this exercise.

We will release our re-processed UI-PRMD skeleton data as well as the annotations together with ExeCheck dataset.

\subsection{Experimental Methodology}

To train and evaluate ExeChecker, we split the ExeCheck dataset and used the extracted skeleton joints of subjects 1--6 for training and those of subject~7 for testing. We used the segmented sequences and augmented them by flipping the left and right sides, resulting in 120 sequences for correct and incorrect movements in the training set and 20 in the testing set per exercise. 
We then exhaustively composed all possible combinations of triplets. Anchors and positives are sampled from correct exercises regardless of orders due to the in-triplet swap strategy, and negatives are sampled from incorrect exercises. As a result, for each exercise in the ExeCheck, $\binom{60}{2} \times \binom{60}{1} = 106,200$ triplets are used for training and $\binom{10}{2} \times \binom{10}{1} = 450$ triplets are used in validation for each exercise.

For the UI-PRMD, we used the skeleton joints of subjects 1--9 of UI-PRMD for training and those of subject 10 for testing, which corresponds to approximately (some movement sequences do not have 10 repetitions) 180 sequences per correct and incorrect exercises in the training set and 20 per correct and incorrect exercises in the testing set after repetition segmentation and mirror augmentation. We then composed triplets in a similar way as we did for the ExeCheck dataset, resulting in approximately 2.8 million ($\binom{180}{2} \times \binom{180}{1} $) triplets for training and 3800 ($\binom{20}{2} \times \binom{20}{1}$) triplets for validation for each exercise.
We randomly/uniformly sample 160 frames and then randomly/centrally crop 128 frames for training/test splits for the ExeCheck dataset and 104 then 88 for the UI-PRMD dataset.
All the models are trained using the AdamW optimizer with a start learning rate of 0.001. We step the learning rate scheduler exponentially at a ratio of 0.9 twice every epoch. All models are trained on a 48G GPU node for a maximum of 10 epochs. We use a batch size of 20 for both datasets.

We use the same configuration as used by Hu \etal~\cite{hu2022spatial} for extracting features with STGAT. We set the dimension of the embedding space for triplets training to 128 for all the models. For calculating the JoA scores, we pull the cross-spacetime attention map of the input sequence, take the average of the 8 heads, use the center frame as suggested by the authors, and sum the attention values each joint received as their raw joint scores.

\subsection{Average JoA Results}

We report the results of a baseline that uses the hop-adjusted Canonical Time Warping (CTW)~\cite{zhou2009canonical}. CTW is an extension of Dynamic Time Warping (DTW) that incorporates canonical correlation analysis as a measure of spatial alignment. Zhou and De La Torre showed that CTW outperforms DTW in human motion alignment~\cite{zhou2009canonical}. 
Following their method, we first aligned the segmented correct (CE) and incorrect exercises (IE). For a fair comparison, we used the normalized sequences as inputs for ExeChecker for the CTW as well. The aligned path is then used for indexing the corresponding time points in CE and IE. We aggregated the Euclidean distance per joint between the aligned CE and IE and considered those distances as the joint's raw scores. A higher score, \ie, a larger distance, suggests that the joint in IE is further away from CE, indicating an incorrect movement.  For JoA evaluation, our method min-max normalizes the raw scores to zero and one. 

In our preliminary studies, we found that terminal joints like hands and feet tend to always have higher values and yield low JoA scores for both ExeCheck and UI-PRMD datasets.
We note that distal joints with higher degrees of freedom compared to proximal joints are biased to higher scores due to the nature of the alignment. To correct for this bias, we adjusted the raw joint score $s^{\text{r}}$ obtained from CTW by its number of hops to the root (pelvis) such that the adjusted score $s_i$ of joint $i$ is equal to 
$s_i^{\text{r}} / (h_i+1)$, with $h_i$ being its hop number.
This adjustment increased the JoA scores for most of the exercises. We use the adjusted CTW joint scores to calculate JoA scores as a baseline and the visualizations.

We report average JoA scores of the 10 exercises in ExeCheck dataset in ~\Cref{tab:joa_exec} and those of UI-PRMD in~\Cref{tab:joa_uipr}.

\begin{table}[th]
    \caption{Average JoA scores per exercise on ExeCheck Dataset. ExeChecker outperforms CTW for 9 of 10 exercises.}
    \centering
\begin{tabular}{l|c|c|c|c|c|c|c|c|c|c}
\textbf{Exercises} &
  \textbf{0-AC} &
  \textbf{1-FL} &
  \textbf{2-HKR} &
  \textbf{3-HA} &
  \textbf{4-LE} &
  \textbf{5-SA} &
  \textbf{6-SER} &
  \textbf{7-SF} &
  \textbf{8-SSS} &
  \textbf{9-Sq} \\ \hline
CTW        & 0.281 & 0.208 & 0.356 & 0.254 & 0.299 & 0.189 & 0.588 & 0.259 & 0.346 & 0.367 \\ \hline
ExeChecker & \textbf{0.429} & \textbf{0.403} & \textbf{0.402} & \textbf{0.633} & \textbf{0.388} & \textbf{0.223} & 0.207 & \textbf{0.373} & \textbf{0.387} & \textbf{0.488} \\ \hline
\end{tabular}
\vspace{0.2cm}
\label{tab:joa_exec}
\end{table}

\begin{table}[th]
    \caption{Average JoA scores per exercise on UI-PRMD~\cite{vakanski2018data}. We used the same order of exercises as in Table~1. ExeChecker outperforms CTW for all but one exercise.}
    \centering
\begin{tabular}{l|c|c|c|c|c|c|c|c|c|c}
\textbf{Exercises} &
  \textbf{m01} &
  \textbf{m02} &
  \textbf{m03} &
  \textbf{m04} &
  \textbf{m05} &
  \textbf{m06} &
  \textbf{m07} &
  \textbf{m08} &
  \textbf{m09} &
  \textbf{m10} \\ \hline
CTW & 0.414 & 0.479 & 0.473 & 0.165 & 0.305 & 0.382 & 0.183 & 0.550 & 0.278 & 0.295 \\ \hline
ExeChecker & \textbf{0.543} & 0.454 & \textbf{0.517} & \textbf{0.517} & \textbf{0.447} & \textbf{0.733} & \textbf{0.539} & \textbf{0.614} & \textbf{0.543} & \textbf{0.664} \\ \hline
\end{tabular}
\vspace{0.2cm}
\label{tab:joa_uipr}
\end{table}

\begin{figure}[tb]
  \centering
  \includegraphics[width=0.9\linewidth]{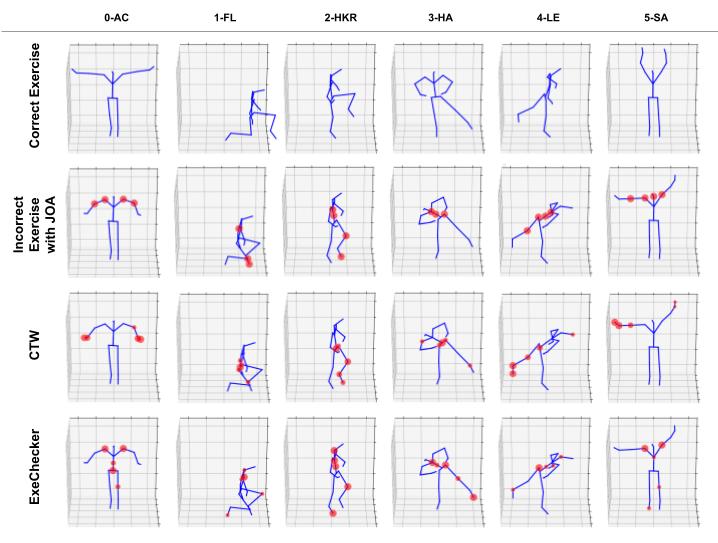}
  \caption{\textbf{Qualitative results on the ExeCheck dataset}.  The first row shows the correct form of each exercise, and the following three rows show an example of the incorrect form per exercise. The skeletal joints in the second row are overlaid with the corresponding JoAs (\Cref{tab:dataset_exec}) shown as red circles of equal sizes.
  For CTW and ExeChecker, we highlight the top 5 joints with the highest attention scores evaluated on the incorrect exercise. The size of the circle is proportional to the computed score. Note that the number of JoAs varies among exercises; hence, the number of red circles in row~2 differs between exercises. 
  The JoAs and their scores identified by ExeChecker (row~4) are better than those identified by CTW (row~3) in matching the expert-defined JoAs shown in row~2.
  }
  \label{fig:exec_result}
\end{figure}

\begin{figure}[tb]
    \centering
    \includegraphics[width=0.9\linewidth]{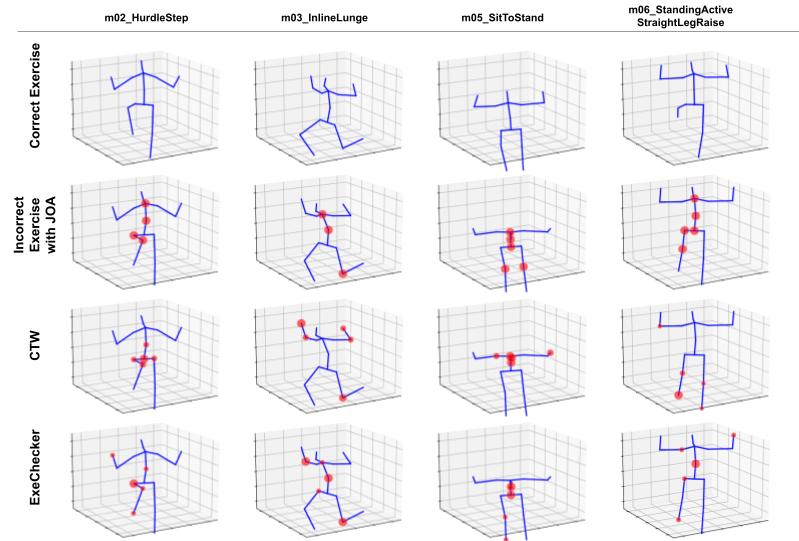}
    \caption{\textbf{Qualitative results on the UI-PRMD} Similar to \Cref{fig:exec_result}, we show and compare our results with CTW and JoA. Compared to the CTW, ExeChecker is good at detecting the torso joint, which is of key importance because it indicates that the subject did not maintain the body trunk upright, a very common mistake for a variety of exercises.
    }
    \label{fig:uipr_result}
\end{figure}

\subsection{Qualitative Results}
We present our qualitative results on the ExeCheck dataset in \Cref{fig:exec_result} and results on the UI-PRMD dataset in \Cref{fig:uipr_result}. Both figures show that ExeChecker in general focuses on the right joints that need attention and such feedback is more informative than the CTW. Specifically, 
as shown in the first column in the~\Cref{fig:exec_result}, 0-AC (Arm Circle), although missing the elbows, ExeChecker correctly focuses on the shoulders with great attention, which is more appropriate than CTW's output; CTW assigns the highest joint scores to the wrists and hands despite the hop-adjustment. The second column of 1-FL (Forward Lunge) presents an interesting case. We list the left foot as one of the joints of attention for the forward lunge because a too-small step is the reason causing the knee to pass the toe, a common error that can put too much stress on the knee joint and could eventually lead to knee injury. Although the ExeChecker did not catch the left foot, it highlighted the left knee as shown in the example. 

The third column, 2-HKR (High Knee Raise) shows that ExeChecker not only highlights the knee, which in this incorrectly performed exercise was not raised to enough height, but also focuses on the joints along the spine (spNavel, spChest, and neck being extra), which is proper feedback for a wobbling upper body. CTW is only able to catch the knee problem but fails to detect the wobbling. 
The column of 3-HA further demonstrated the strength of ExeChecker in detecting the tilting of the upper body with additional attention to the leg in motion. The column of 4-LE shows that ExeChecker pays the heaviest attention to the correct hip that is responsible for the leg extension. It still catches the upper body leaning forward problem in the top 5 but with less attention. 

The following column of 5-SA shows that ExeChecker correctly pays the highest attention to the shoulders; it then assigns attention to the chest and lower body instead of the neck and the elbows though.
Overall, ExeChecker is able to highlight the subtle but err-leading movements like hunching shoulders and unstable upper body, which are usually missed by CTW, which is based on pairwise alignment. 
For the plots of exercises 6--9 of ExeCheck, please refer to Supplementary Figure 2.

Qualitative results by models trained on UI-PRMD are shown in \Cref{fig:uipr_result}. In the first column, the incorrect hurdle step features an non-neutral femur position caused by the right hip and a slightly bent body trunk. ExeChecker highlights the hip with the greatest attention and then the leg (knee, ankle) and torso. In comparison, CTW assigns the greatest joint score to the pelvis. In the second column, the inline lunge, both ExeChecker and CTW highlight the rear knee, which should not touch the ground. However, CTW does not catch the torso, which should be highlighted because the subject did not maintain an upright trunk posture.
The third column is about sit-to-stand. The subject incorrectly raised the pelvis while standing up. CTW catches the torso and neck but misses the pelvis. It instead assigns a score to the head. ExeChecker on the other hand, misses the neck but it correctly focuses on the pelvis, which is the more important one. 
In addition, the subject here had a slight knee valgus collapse on the right side and ExeChecker pays moderate attention to that knee.
In the last column, straight leg raise, the incorrect movement is about the body trunk not being upright, deviated 
pelvis, and insufficient right hip flexion, which leads to not raising the leg high enough. CTW misses the torso again and focuses mostly on the right ankle and then the knee. On the other hand, ExeChecker notices the hip after assigning most of the attention to the torso.

\subsection{Ablation Studies}

We investigated the effects of different distance functions used in the triplet loss, which revealed that no distance function is a clear winner for all exercises (\Cref{tab:aba}).

We also investigated the effects of bone orientations and report our results in the same table. We decouple the skeleton and each joint essentially is replaced by the bone vector from its parent to itself. That is,  joint $J_i$ instead of $(x_i,y_i,z_i)$ is now represented by ($\delta x_i$, $\delta y_i$, $\delta z_i$), where $(\delta x_i, \delta y_i,\delta z_i) = (x_i,y_i,z_i) - (x_{f(i)},y_{f(i)},z_{f(i)})$, $f(\cdot)$ denotes a functional relationship and $f(i)$ is the parent joint of $J_i$ following the kinematic chain. The root joint, pelvis ($J_0$) has $(\delta x_0,\delta y_0,\delta z_0) = (0,0,0)$. 
In general, the decoupled sequence shows degraded JoA scores possibly due to the information loss when subtracting the parent joints. Arguably, the cosine similarity works better with the decoupled sequences.

\begin{table}[th]
    \caption{Average JoA scores per exercise on ExeCheck Dataset. ExeChecker uses Euclidean distances for computing triplet loss. EC-cos use cosine similarity for computing the pairwise distances between (anchor, positive) and (anchor, negative) pairs. Row 3 and 4 use bone vectors as the model inputs with EC-Bone using Euclidean distances and EC-Bone-cos using cosine similarities as their distance metric.}
    \centering
\begin{tabular}{l|c|c|c|c|c|c|c|c|c|c}
\textbf{Exercises} &
  \textbf{0-AC} &
  \textbf{1-FL} &
  \textbf{2-HKR} &
  \textbf{3-HA} &
  \textbf{4-LE} &
  \textbf{5-SA} &
  \textbf{6-SER} &
  \textbf{7-SF} &
  \textbf{8-SSS} &
  \textbf{9-Sq} \\ \hline
ExeChecker & 0.429 & 0.403 & 0.402 & 0.633 & 0.388 & 0.223 & 0.207 & 0.373 & 0.387 & 0.488 \\ 
EC-cos     & 0.295 & 0.397 & 0.638 & 0.274 & 0.444 & 0.674 & 0.591 & 0.377 & 0.588 & 0.608 \\ \hline
EC-Bone       & 0.031 & 0.493 & 0.038 & 0.145 & 0     & 0.536 & 0 & 0.343 & 0.114 & 0.004 \\
EC-Bone-cos   & 0.123 & 0.376 & 0.034 & 0.439 & 0.250 & 0.814 & 0.717 & 0.540 & 0.513 & 0.154 \\ \hline
\end{tabular}
\vspace{0.2cm}
\label{tab:aba}
\end{table}

\section{Conclusions}

In this paper, we presented ExeChecker, a framework for identifying and visualizing the joints of attention in incorrectly conducted exercises. We further proposed a JoA scoring method to quantitatively evaluate the identification. Both quantitative and qualitative results show that ExeChecker in general focuses on the right joints. Benefiting from contrastive learning using triplets, ExeChecker can detect small yet crucial joint movements that make the exercise incorrect, like shoulder joints for struggling shoulders, and spine joints for wobbling, leaning, or tilting of the upper body. Those small movements are usually missed by the baseline method CTW.
Importantly, the composition of the triplets is based on the paired correct and incorrect sequences, provided by our ExeCheck dataset.  As to the best of our knowledge, ExeCheck and UI-PRMD are the only two datasets that have paired movement sequences. Moreover, in addition to skeletal sequences, ExeCheck also provides RGB-D videos.
A limitation of the ExeCheck dataset is that it only includes a set of common mistakes and the JoA labels are
specific to those mistakes. Future work could expand the dataset by adding incorrect variants and annotating the corresponding JoAs.  A user study is planned that analyzes how helpful the visual feedback showing "Where did I go wrong?" is for exercising participants, particularly Parkinson's patients.  We plan to incorporate the ExeChecker visualization into the ExerciseCheck platform~\cite{pandit2019exercisecheck} for this user study.

\section*{Acknowledgements}

We thank the participants of our data collection study, and 
Mona Jalal and Shreya Pandit for helping with the data acquisition. 
We are grateful to Timothy Nordahl, Doctor of Physical Therapy,  for his instructions on the correct performance of rehabilitation exercises in our study, and for his advice on how our participants should move to illustrate typical incorrect performance of these exercises.   
We thank Professor Aleksandar Vakanski, University of Idaho, for kindly providing us access to the original videos of the UI-PRMD dataset.


%
%
\bibliographystyle{splncs04}
\bibliography{humanpose_aqa}

\end{document}